\documentclass[letterpaper, 10 pt, conference]{ieeeconf}  
\IEEEoverridecommandlockouts                              

\usepackage{graphics} 
\usepackage{epsfig} 
\usepackage{stfloats}
\usepackage{times} 
\usepackage{amsmath} 
\usepackage{amssymb}  
\usepackage[hang]{subfigure} 
\usepackage{cite}
\usepackage{siunitx}
\usepackage{soul}
\usepackage{color}
\usepackage{xcolor}
\usepackage[ruled,linesnumbered]{algorithm2e}
\usepackage{caption}

\soulregister\ref7  
\soulregister\cite7 
\soulregister\eqref7
\soulregister\prettyref7

\usepackage{multirow}
\usepackage{algpseudocode}
\usepackage{makecell}
\usepackage{booktabs}
\usepackage{threeparttable}

\bstctlcite{IEEEexample:BSTcontrol}

\usepackage{prettyref}
\newrefformat{fig}{Fig.~\ref{#1}}
\newrefformat{sec}{Sec.~\ref{#1}}
\newrefformat{tab}{Tab.~\ref{#1}}
\newrefformat{algo}{Algo.~\ref{#1}}
\newrefformat{eq}{(\ref{#1})}

\usepackage{color} 

\title{\LARGE \bf
Reinforcement Learning with Generalizable Gaussian Splatting}
\author{Jiaxu Wang$^{1,*}$, Qiang Zhang$^{1,2,*}$, Jingkai Sun$^{1}$, Jiahang Cao$^{1}$,  Gang Han$^{2}$,  Wen Zhao$^{2}$,\\ Weining Zhang$^{2}$, Yecheng Shao$^{3, 4}$, Yijie Guo$^{2}$, Renjing Xu$^{1, \dagger}$
\thanks{$^{*}$ are equal contributors, $^{\dagger}$ are the corresponding authors}
\thanks{$^{1}$The authors are with The Hong Kong University of Science and Technology (Guangzhou), China.
{\tt\small  \{jwang457, qzhang749, jsun444, \}@connect.hkust-gz.edu.cn, renjingxu@ust.hk} }
\thanks{{$^{2}$The authors are with Beijing Innovation Center of Humanoid Robotics Co., Ltd.}
{\tt\small  jack.guo@x-humanoid.com}}
\thanks{$^{3}$The author is with Center for X-Mechanics, Zhejiang University, China.}
\thanks{$^{4}$The author is with Institute of Applied Mechanics, Zhejiang University, China.
 \tt\small shaoyecheng@zju.edu.cn  }%
}

\begin{document}
\maketitle
\thispagestyle{empty}
\pagestyle{empty}

\begin{abstract}
An excellent representation is crucial for reinforcement learning (RL) performance, especially in vision-based reinforcement learning tasks. The quality of the environment representation directly influences the achievement of the learning task. 
Previous vision-based RL typically uses explicit or implicit ways to represent environments, such as images, points, voxels, and neural radiance fields. However, these representations contain several drawbacks. They cannot either describe complex local geometries or generalize well to unseen scenes, or require precise foreground masks. 
Moreover, these implicit neural representations are akin to a ``black box", significantly hindering interpretability.
3D Gaussian Splatting (3DGS), with its explicit scene representation and differentiable rendering nature, is considered a revolutionary change for reconstruction and representation methods.
In this paper, we propose a novel Generalizable Gaussian Splatting framework to be the representation of RL tasks, called GSRL. Through validation in the RoboMimic environment, our method achieves better results than other baselines in multiple tasks, improving the performance by 10$\%$, 44$\%$, and 15$\%$ compared with baselines on the hardest task. 
This work is the first attempt to leverage generalizable 3DGS as a representation for RL.
\end{abstract}
\section{Introduction}
\label{sec:intro}
In reinforcement learning (RL), obtaining a high-quality representation is crucial for problem-solving~\cite{finn2016deep,dwibedi2018learning,vecerik2021s3k,jonschkowski2017pves,kulkarni2019unsupervised,laskin2020curl,manuelli2020keypoints}. This challenge becomes even more pronounced in vision-based RL tasks, where the ability to derive effective representations from complex visual scenes is essential for developing successful downstream strategies. Particularly in application scenarios like robotic manipulation, high-quality environmental representations would influence the success rate of tasks. To tackle this issue, we extract information from high-dimensional visual data and convert it into representations that are compatible with deep RL algorithms. This process demands not only the skill to comprehend and process high-dimensional data but also the ability to capture the fundamental characteristics of the environment. Such capabilities allow robots or other automated systems to make precise decisions and actions based on these characteristics. Consequently, developing a versatile representation method that can accurately capture environmental features.

In previous research, representation methods in RL have primarily been divided into two types: low-dimensional and high-dimensional representations. Low-dimensional representation methods focus on expressing environmental information in a structured form, such as object positions, postures, or even shapes from visual inputs~\cite{jonschkowski2017pves}. Even though low-dimensional representation methods can explicitly and accurately express environmental information, they are impractical to obtain. Conversely, high-dimensional representation methods tend to adopt an end-to-end approach to handle high-dimensional visual information. This method directly transforms visual information into high-dimensional features by using pre-trained models, such as~\cite{he2016deep, liu2021swin, vaswani2017attention}.

However, only 2D images are not capable of perceiving 3D real-world structures. Thus researchers adopt 3D-aware explicit representations in RL, such as RGB-D images \cite{sun2022rgbdrl, gao2022fras}, multiview images\cite{fan2022dribo, yang2022self}, voxels \cite{ze2023visual}, and point clouds \cite{ling2023pointrl, qin2023dexpoint}. Unfortunately, these explicit scene representations cannot describe complex 3D local geometries due to the limitation of their resolutions, therefore, they cannot be considered an expected representation for RL.
With the advancement of the Neural Radiance Field (NeRF) \cite{mildenhall2021nerf, wang2023learning, barron2022mip,wang2024learning,tancik2022block}, which is a novel implicit, 3D-consistent scene representation. 
Researchers have applied such implicit NeRF formats to environmental representation~\cite{driess2022reinforcement,ze2023gnfactor, shim2023snerl}. Nevertheless, these NeRF-based methods either require high-quality foreground masks to distinguish the target objects or need high-level semantic feature maps produced by other large-scale deep learning models such as stable diffusion. More importantly, this type of method cannot smoothly generalize to unseen scenes because they tend to use a single vector to represent the whole scene. Therefore, an important research question arises: can we develop a new environmental representation that can both explicitly express 3D-aware structural information and capture 3D-consistent local geometry?

\begin{figure*}[t]
	\setlength{\tabcolsep}{1.0pt}
	\centering
	\begin{tabular}{c}
		\includegraphics[width=0.9\textwidth]{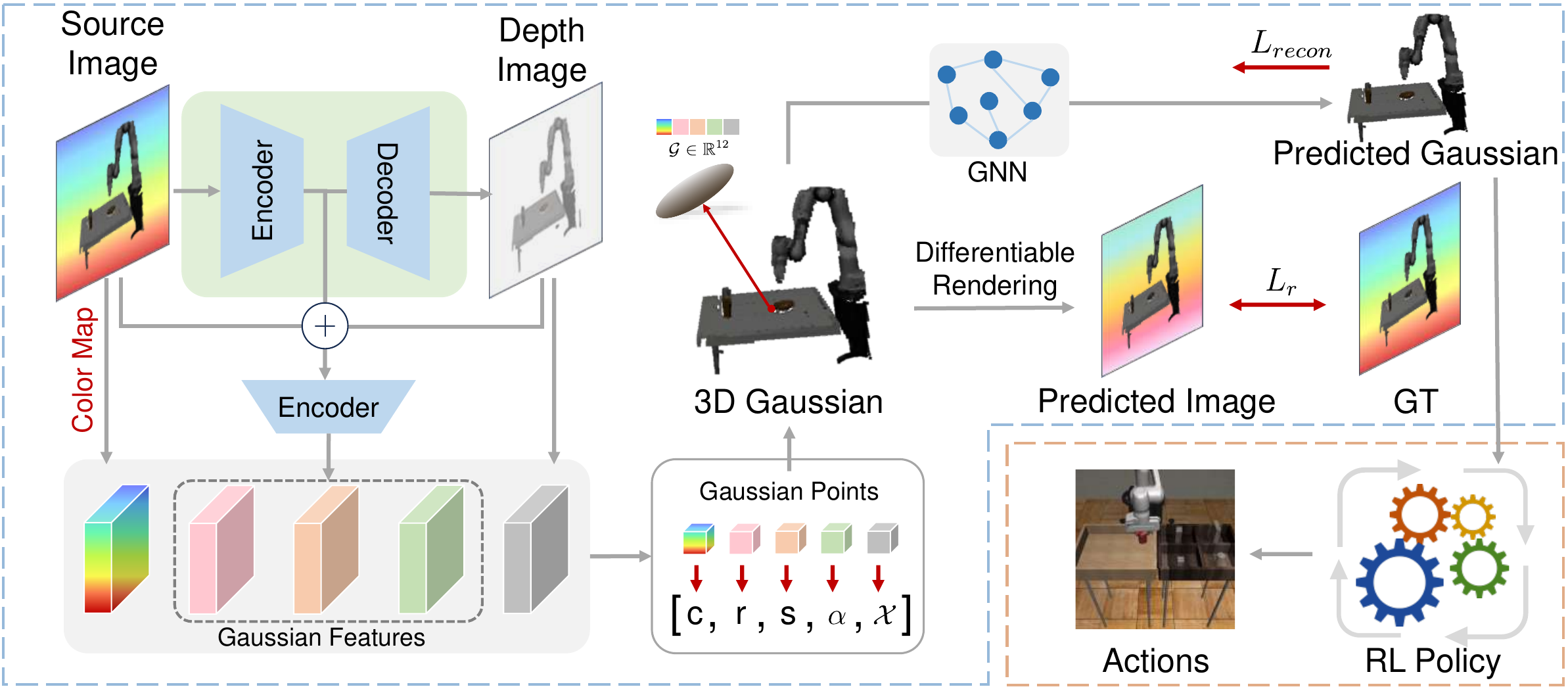} \\
	\end{tabular}
	\caption{The overview of the main pipeline. Contents in the blue dashed line represent the training of the generalizable Gaussian prediction module. This module converts image observation into a 3D-consistent and geometry-aware 3D Gaussian cloud. Contents in the orange dashed line denote the RL training module, which is fed with the reconstructed 3D Gaussians to predict the target actions. }
	\label{fig:main-fig}
	\vspace{-0.45cm}
\end{figure*}

3D Gaussian Splatting (3DGS)~\cite{kerbl20233d} provides an approach to answer the question. This technique not only enables 3D-consistent content expression but also explicitly represents detailed local geometries. Owing to its differentiable properties, it can be integrated into many deep learning frameworks. However, conventional 3DGS requires per-scene optimization that obstacles its usage in RL. 
Motivated by this, we introduce a generalizable 3DGS framework that is pretrained for obtaining 3D-aware prior knowledge of specific tasks, enabling the extraction of high-quality 3DGS from the visual observations for RL tasks. We select RoboMimic~\cite{robomimic2021} as our training and evaluation environment. RoboMimic is a sophisticated dataset and benchmark for robotics research. To put it in a nutshell, our contributions can be summarized as follows:
\begin{enumerate}
    \item To the best of our knowledge, we propose to adopt the pretrained generalizable 3D Gaussian as a representation for RL for the first time. Our approach is illustrated in Fig.~\ref{fig:main-fig}. This innovation not only introduces an efficient way of environmental representation to the RL field but also broadens the application prospects of 3DGS technology in various RL tasks.
    \item Our approach has demonstrated outstanding performance by validating our method across multiple tasks in the RoboMimic benchmark. This achievement not only proves the effectiveness of our proposed representation method but also provides new insights for future vision-based RL.
\end{enumerate}


\section{Related work}
\label{sec:related}

\subsection{3D Scene Representation: Implicit and Explicit Way}
Traditional methods that directly optimize explicit 3D geometries, such as mesh \cite{liuGeneralDifferentiableMesh2020}, voxel\cite{ sitzmannDeepvoxelsLearningPersistent2019} and point cloud\cite{qi2017pointnet}. These representations use unified primitives to describe 3D structures. However, they cannot effectively describe detailed local geometries due to the limited resolutions. Recently the use of networks for implicitly representing scenes has prevailed \cite{sitzmannDeepvoxelsLearningPersistent2019, tancikFourierFeaturesLet2020}. Among them, the Neural Radiance Field (NeRF)\cite{mildenhall2021nerf} attracts the most attention.  Recent studies have employed NeRF for various purposes, including dynamic reconstruction \cite{yan2023nerf, cao2023hexplane}, physical reasoning \cite{wang2024physical}, and reinforcement learning \cite{ze2023gnfactor, shim2023snerl, driess2022reinforcement}. More recently, \cite{kerbl20233d} proposed the 3D Gaussian Splatting to realize a remarkable rendering speed and high-quality 2D images by combining the advantages of both implicit and explicit representations. 

\subsection{Learning Representations for Reinforcement Learning}
In the realm of RL, environments are typically described using intuitive, explicit methods, such as detailing the position~\cite{du2021vision,labbe2020cosypose,sundermeyer2021contact}, posture~\cite{wang2019densefusion,he2021ffb6d}, and motion state~\cite{jonschkowski2017pves} of objects. However, these techniques have two significant challenges: first, in complex scenarios—such as complex shapes or soft materials—it becomes challenging to form structured representations; second, in real-world applications, these representations cannot be directly obtained from the physical world. 
With the progress in computer vision, researchers employ vision-based representations to train RL, such as \cite{sun2022rgbdrl, gao2022fras, fan2022dribo, yang2022self}. However, 2D visions hardly reflect the real structures of objects because we live in the 3D world. Therefore, researchers investigate 3D-based representations including explicit and implicit ones. For the explicit ones, \cite{ze2023visual} train RL with voxel-based representation. \cite{ling2023pointrl} and \cite{qin2023dexpoint} propose the benchmarks of point-based RL learning. \cite{ze2023gnfactor}, \cite{shim2023snerl}, and \cite{driess2022reinforcement} propose to adopt implicit NeRF for scene modeling, encapsulating scene information within implicit vector features. Nonetheless, methods based on NeRF generally exhibit limited generalization to unseen scenes and usually require foreground masks. 

3DGS~\cite{kerbl20233d} inherits from the point cloud, but with per-point geometry features to describe more detailed local structures. However traditional 3DGS does not match the requirement of RL. In this paper, we introduce a pretrained generalizable 3DGS pipeline for addressing this issue.

\section{Preliminary}
\label{sec:preliminary}
\subsection{Reinforcement Learning}
Our RL method is conceptualized within the framework of Markov Decision Process (MDP). This MDP is described by a tuple $(\mathcal{S},\mathcal{A},\mathcal{R},p,\gamma)$, where $\mathcal{S}$ denotes the state space, $\mathcal{A}$ signifies the action space, $\mathcal{R}$ represents the reward function, $p$ delineates the transition probabilities from the current state to the subsequent state, and $\gamma \in [0,1]$ stands for the discount factor applied to rewards. At each time step $t$, the agent interacts with the environment by receiving an observation. Subsequently, the agent outputs an action $a_t \in \mathcal{A}$ based on a policy $\pi(a_t|s_t)$. According to the action, the state of the robot transitions from $s_t$ to $s_{t+1}$ based on the transition function $s_{t+1} \sim p(s_{t+1}|s_t,a_t)$. Additionally, the agent receives a reward $r_t=\mathcal{R}(s_t,a_t)$ at each time step. The objective is to maximize the return, achieved by optimizing the parameters $\theta$ of the policy:
\begin{equation}
    \textnormal{arg}\max_{\theta} \mathbb{E}_{(s_t,a_t) \sim p_\theta(s_t,a_t)} \left[ \sum_{t=0}^{T-1} \gamma^t r_t\right]
\end{equation}
where $T$ denotes the time horizon of MDP. 
\subsection{3D Gaussian Splatting}
3DGS parameterize a 3D scene as 3D Gaussian primitives, each of primitives includes a mean ($\mu_k$), a covariance ($\sum_k$), an opacity ($o_k$) and spherical harmonics coefficients ($\textbf{SH}_k$) that represents the color information. These primitives can be rendered and accessed to produce novel views via Gaussian rasterization. To facilitate optimization by backpropagation, the covariance matrix can be decomposed into a rotation matrix ($\textbf{R}$) and a scaling matrix ($\textbf{S}$):
\begin{equation}
    \Sigma = RSS^TR^T
    \vspace{-1mm}
\end{equation}
The camera parameters and poses can be estimated via structure from motion such as Colmap, the projection of the 3D Gaussian to 2D image plane can be transformed by the view transformation ($\textbf{W}$) and the projection transformation. However, the projection matrix usually destroys the shape of Gaussian primitives due to its nonlinearity. To solve this issue, the Jacobian of the affine approximation $\textbf{J}$ of the projective transformation is applied, as in:
\begin{equation}
    \Sigma^{'} = JW\Sigma W^TJ^T
    \vspace{-1mm}
\end{equation}
where the $\Sigma^{'}$ is the projected 2D covariance. After all Gaussians are transformed to the 2D planes, the final pixel color can be obtained by $\alpha$-blend.
\begin{equation}
    C = \sum_{i\in \mathcal{N}}c_i\alpha_i\prod_{j=1}^{i-1}(1-\alpha_j)
    \vspace{-1mm}
\end{equation}
in which $c_i$ is computed from the spherical harmonics coefficients \textbf{SH}. $\alpha_{i}$ denotes the soft occupation Gaussian point $i$ at 2D space, which can be calculated by $\alpha_i(x) = o_iexp(-\frac{1}{2}(x-\mu_i)^T{\Sigma}_{i}^{T}(x-\mu_i))$.  In this work, we replace the \textbf{SH} with a single RGB color vector to facilitate simplicity. 
\vspace{-2mm}
\section{Method}
\label{sec:method}
Conventional 3DGS requires per-scene optimization from multiview images to adjust the Gaussian properties initialized randomly, therefore it can not be utilized as the state representation of RL. This is because it is impractical and extremely time-consuming to optimize a 3D Gaussian model at each moment of getting observations. Therefore, we propose to directly estimate 3D Gaussian clouds from multi-view images in a generalizable way. The intuition is that the scenes and observations for a certain RL task are similar, therefore they share similar mappings from 2D images to 3D local geometries, and those local geometries can be seamlessly described by the properties of 3D Gaussians, i.e. the shape of each Gaussian point. Hence, the generalizable GS framework inherently learns the prior knowledge to map 2D patches to 3D local structures. In this case, RL algorithms can directly operate on this representation. The main pipeline of the method is demonstrated in Fig.~\ref{fig:main-fig}.

This image-conditioned generalizable Gaussian representation has two-fold advantages. On the one hand, 3D Gaussian inherits the nature of the point cloud. Furthermore, the properties of 3D Gaussian allow more detailed descriptions of 3D local geometry, thus producing better geometry-aware feature representation. On the other hand, 3D Gaussians could be 3D-consistent because they are constructed from multiview images, thereby being robust to occlusions and generating 3D-consistent features. There are two steps in our framework. First, we train the image-conditioned 3DGS estimator network. This network can predict a 3D Gaussian cloud from given single or multiple images. In addition, we adopt a GNN network as a smoothness regularizer to enhance the 3D consistency. Second, we integrate and freeze the pretrained Gaussian estimator into the RL environment to convert the output observation of the environment into 3D Gaussians in real-time on which the RL policy is trained. 

\subsection{Generalizable GS framework}
This section introduces the pretrained 3D Gaussian encoder. Given $N$ images to describe a scene $\{I_n \in R^{H\times W\times 3}\}_1^N$ and their corresponding camera parameters $\{C_n=\{K_n, P_n\}\}_1^N$, we aim to reconstruct the 3D Gaussian representation conditioned by the given images, and the reconstructed Gaussian cloud can be accessed to render novel images with arbitrary viewpoints. We divide this image-conditioned Gaussian encoder into three main components: a depth estimator module, a Gaussian regressor module, and a Gaussian refinement module. In the training paradigm, when an image is randomly selected from the dataset (marked as $I_t$), we intentionally select the two closest nearby images of it ($I_{s1}$ and $I_{s2}$). Then the two source views synthesize the target view used to optimize all networks via the difference with the $I_t$, which can be described as follows:
\begin{equation}
    I_t^{'}=\mathcal{R}(\mathcal{G}(I_{s1}, I_{s2})|K_t, P_t)
    \label{eq: main pipeline equation}
\end{equation}
where $\mathcal{R}$ denotes the rendering function of 3DGS, which is inspired by the original \cite{zheng2024gps}, $\mathcal{G}$ refers to the 3D Gaussian encoder. $K_t, P_t$ are the camera intrinsic and pose of target view $t$, and $I_t{'}$ is the predicted target view via 3DGS. In the depth estimation module, we predict the absolute depth value for each pixel to transform a 2D image grid into a 3D coordinate space. The 3D coordinate of each point is regarded as the $\mu$ in the Gaussian properties. Then the Gaussian regressor predicts the rest of the Gaussian properties in a pixel-wise manner, which are transformed to 3D space accompanied by the depth value. Last, to improve the consistency of features, we define the Gaussian refinement operation to smooth the features in 3D space. 

\noindent\textbf{Depth Estimation}. Estimating the depth map is crucial to bridge 2D images and 3D Gaussians. It is noted that the input of this module is a pair of stereo images. Therefore, depth prediction is equivalent to disparity prediction, and any alternative depth estimation theories can be adapted to this step. First, the two source views are fed to two extractors to extract semantic features.
\begin{equation}
    F_{s1}, F_{s2} = Ex(I_{s1}), Ex(I_{s2})
\label{eq : image feature extraction}
\end{equation}
$Ex$ refers to the feature extractor and the two networks share the same parameter. The extracted features are used to build cost volume by homography transformation \cite{yao2018mvsnet}. The disparity value for each pixel is predicted through the cost volume. In practice, our model predicts the normalized log disparity map. The eventual absolute depth value can be converted from the prediction via:
\begin{equation}
    D_{pred}=exp(D_{max} \cdot \sigma(\mathcal{D}(F_{s1}, F_{s2})))
    \label{eq : depth prediction}
\end{equation}
Here $\mathcal{D}$ refers to the disparity prediction network that we introduced above. $\sigma$ is the Sigmoid activation. $D_{max}$ indicates the maximum value of disparity across the dataset. We swap $I_{s1}$ and $I_{s2}$ in Eq.~\ref{eq : depth prediction} to obtain all their per-pixel depth. Notably, we also train another depth predictor network with single image input because in some experiments it only allows one image as the observation, and the results show minimal discrepancy compared to the paired image input. When the camera set includes a global camera and a robot hand camera, we train two depth predictors for the two cameras separately. 

\noindent\textbf{Gaussian Properties Prediction}. Each 3D Gaussian is parameterized by five independent properties to define its shape and appearance, i.e. $G=\{\mathbf{X}, \mathbf{R}, \mathbf{S}, \mathbf{c}, o\}$. As we stated before, the color attribute in conventional Gaussians is defined by spherical harmonics, but in this work, we replace it with the RGB vector to reutilize the image pixel color, thereby $\mathbf{c}=I_s$. 
Furthermore, we can obtain the $\mathbf{X}$ from the predicted depth map via:
\begin{equation}
    \mathbf{X} = P_t \cdot D_{pred} \cdot K^{-1}_t \cdot \mathbf{u}
    \label{eq. unprojection depth}
\end{equation}
where $\mathbf{u}=\{u,v,1\}$ is homogeneous 2D plane grid coordinates. The other symbols remain the same meaning as the previous content. Hence, this Gaussian regressor module aims to predict the rest properties in a per-pixel manner. We reutilize the extracted feature in the previous module and concatenate it with the source image and predicted depth to obtain a fused feature map.
\begin{equation}
    \mathcal{F}_R = E_{\phi}(D_s \oplus F_s \oplus I_s)
\label{eq. regressor}
\end{equation}
in which $E_{\phi}$ is a UNet-like encoder parameterized by $\phi$, $\mathcal{F}_R \in R^{H \times W \times D_R}$ denotes the fused feature map which is in the full image resolution. Then the $F_R$ is sent to different prediction heads to regress corresponding Gaussian properties. 
\begin{equation}
\textbf{R},~\textbf{S},~o = norm(\mathcal{H}_r(f_R)),~exp(\mathcal{H}_s(f_R)),~ \sigma(\mathcal{H}_o(f_R)) 
\label{sec. regression gs param}
\end{equation}
$\mathcal{H}_r$, $\mathcal{H}s$, and $\mathcal{H}_o$ represent the corresponding decoder heads for the Gaussian parameters, which are formulated by three full convolutional layers with $1\times 1$ convolution kernel. The predicted parameter maps have the same spatial resolution as the source image, i.e. $\textbf{R} \in R^{H \times W \times 4}, \textbf{S} \in R^{H \times W \times 3}, \alpha \in R^{H \times W \times 1}$. Different functions activate each parameter map. $norm$ indicates the normalization function along the channel dimension. $exp$ is the exponential activation. 

\noindent\textbf{Gaussian Refinement}. After obtaining those properties, we can directly render novel views by Gaussian rasterization. However, due to inherent biases in the image, such as color biases related to different view directions, this might lead to some inconsistent noise. To address this issue, we adopt a GNN-based Autoencoder architecture to smooth these Gaussian properties and filter out inconsistent noises. This procedure is formulated as follows.
\begin{equation}
    \mathbf{R^{'}}, \mathbf{S^{'}}, \mathbf{c^{'}}, o^{'}=D_\theta(E_\theta(\mathbf{R}, \mathbf{S}, \mathbf{c}, o|\mathbf{X}))
    \label{eq: auto-encoder}
\end{equation}
We implement the encoder $E_\theta$ and the decoder $D_\theta$ as graph-based networks, similar to~\cite{yang2018foldingnet}. We select the KNN neighbor number $K=16$ to construct subgraphs. The encoder contains 3 MLPs and outputs the graph with 128-dimensional node features. The decoder also includes 3 layers to restore their respective original parameters. The GNN can be considered a regularization term to alleviate high-frequency noises caused by the discrepancy between neighboring views due to its smooth nature. After this operation, we obtain the smoothed 3D Gaussian representation which can be used to render the target view by Eq.~\ref{eq: main pipeline equation}
\vspace{-2mm}
\subsection{Training Strategy}
\vspace{-1mm}
We first pretrain the depth estimation module by using the $L1$ loss function. After the depth estimator converges sufficiently, we freeze it and train the Gaussian regressor and refinement module jointly. The following losses that guide the second training stage are formulated in the following.
\begin{equation}
\begin{aligned}
    &L_{r} = ||\mathcal{R}_{\theta, \phi}(I_{s1,2}|C_t)-I_t||_2^2 \\
    &L_{recon} = ||D_\theta E_\theta(G)-G||^2_2 \\
    &L_{total} = L_r + \lambda L_{recon}
\label{eq: losses}
\end{aligned}
\end{equation}
$L_r$ denotes the rendering loss. $\mathcal{R}_{\theta, \phi}$ includes the Gaussian rendering and all learnable modules introduced above. $I_t$ and $C_t$ are the target image and its corresponding camera parameters. $L_{recon}$ is an auxiliary loss to supervise the reconstruction of Eq.~\ref{eq: auto-encoder}. $\lambda$ refers to the coefficient to balance the two items and we experimentally set it to 0.15.

\begin{figure*}[t]
	\setlength{\tabcolsep}{1.0pt}
	\centering
	\begin{tabular}{c}
		
		\includegraphics[width=0.9\textwidth]{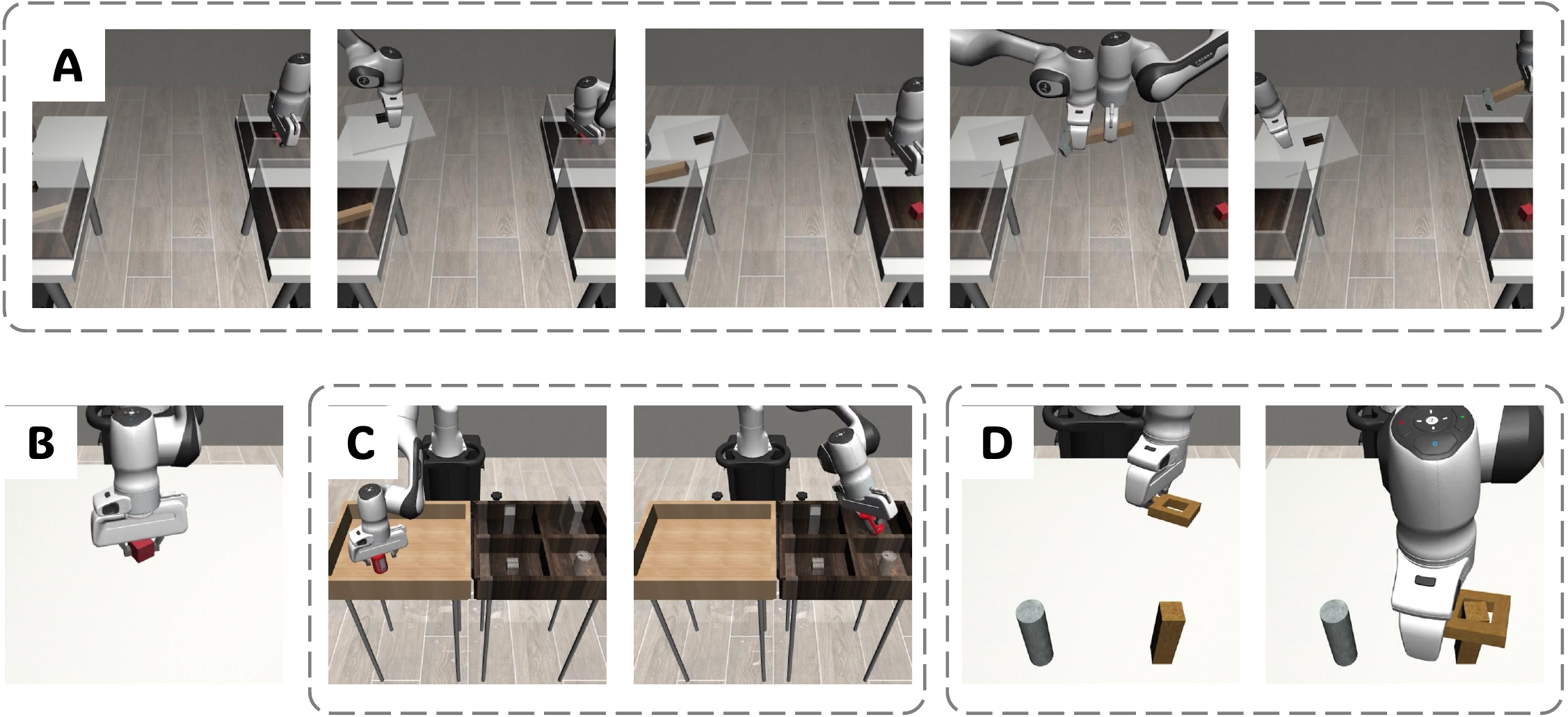} \\
		
	\end{tabular}
	\caption{We evaluate our method in four tasks. \textbf{A} is Transport, meaning two robot arms must take the red box into the target container and transport the hammer to the opposite collaboratively. \textbf{B} is Lift, which forces the arm to take the red box up. \textbf{C} is Can. In this task, the robot arm needs to take the can into a specific area showing the can symbol. \textbf{D} is Square task designed for placing the hollow square object with a handle on the square pillar.  } 
	\label{fig:gstasks}
\end{figure*}
\setlength{\tabcolsep}{3pt}
\begin{table*}[t]
    \centering
        \begin{tabular}{ccccccccccccc}
            \toprule
            \renewcommand{\arraystretch}{0.1}
            \multirow{2}{*}{Representation} & \multicolumn{3}{c}{Lift} & \multicolumn{3}{c}{Can} & \multicolumn{3}{c}{Square} & \multicolumn{3}{c}{Transport}\\
            \cmidrule(lr){2-4} \cmidrule(lr){5-7} \cmidrule(lr){8-10}\cmidrule(lr){11-13}
             & BCQ & IQL & IRIS & BCQ & IQL & IRIS & BCQ & IQL & IRIS & BCQ & IQL & IRIS\\
            \midrule
             \multicolumn{1}{c}{Image}                   & 98.0$\pm$2.0  & 97.0$\pm$1.0                   & 100.0$\pm$0.0           & 71.0$\pm$2.0    & 75.0$\pm$2.0    & 98.0$\pm$0.0     & 42.0$\pm$6.0    & \textbf{4.0$\pm$2.0}    & \textbf{70.0$\pm$3.0}  & 0.0$\pm$0.0      & 0.0$\pm$0.0    & 33.0$\pm$3.0\\
             \multicolumn{1}{c}{Point}                   & 97.5$\pm$1.5  & \textbf{99.0$\pm$1.0}          & 100.0$\pm$0.0           & 68.5$\pm$4.5 & 75.0$\pm$1.0    & 97.5$\pm$1.5 & 28.0$\pm$4.0    & 0.0$\pm$0.0    & 58.5$\pm$10.5  & 0.0$\pm$0.0 & 0.0$\pm$0.0    & 25.0$\pm$5.0\\
             \multicolumn{1}{c}{Voxels}                  & 97.5$\pm$1.5  & 98.5$\pm$1.5                     & 99.5$\pm$0.5            & \textbf{71.5$\pm$0.5} & 73.0$\pm$3    & 98.5$\pm$0.5 & 42.0$\pm$6.0    & 0.0$\pm$0.0    & 69.0$\pm$3.0  & 0.0$\pm$0.0 & 0.0$\pm$0.0    & 31.5$\pm$2.5\\
            \multicolumn{1}{c}{\textbf{Gaussians(ours)}} & \textbf{98.5$\pm$1.5}  & 99.0$\pm$1.0 & \textbf{100.0$\pm$0.0}  & 70.0$\pm$2.0    & \textbf{76.5$\pm$1.5}& \textbf{98.5$\pm$0.5} & \textbf{48.5$\pm$3.5}& 0.0$\pm$0.0    & 70.0$\pm$3.0  & 0.0$\pm$0.0      & 0.0$\pm$0.0    & \textbf{36.0$\pm$2.0}\\
             \bottomrule
        \end{tabular}%
    \caption{Quantitative Comparisons between different visual observation modalities on four tasks with three RL algorithms.}
    \label{tab: result 1}
    \vspace{-5mm}
\end{table*}

\vspace{-1mm}
\section{Experiments}
\vspace{-2mm}
\label{sec:experiments}
We evaluate the proposed novel scene representation on the robomimic \cite{robomimic2021} robot learning platform on various environments and reinforcement learning methods. Robomimic is a framework that can learn RL policy from demonstrations. We select four tasks, namely Lift, Can, Square, and Transport, which are illustrated in Figure~\ref{fig:gstasks}, and three Offline RL algorithms BCQ~\cite{fujimoto2019off}, IQL~\cite{kostrikov2021offline}, and IRIS~\cite{mandlekar2020iris} with four different vision modalities including images, point clouds, voxels, and Gaussian points. All the demonstration data are claimed to be collected by a Proficient Human operator with 200 successful trajectories. The generalizable Gaussian module is trained with some image-depth pairs and their associated camera parameters rendered by the robosuite simulator~\cite{zhu2020robosuite}. 

\begin{table}[t]
    \centering
\scalebox{1.2}{
\begin{tabular}{ccccc}
\toprule
Sample number                 & 2048 & 4096 & 8192 \\ \midrule
Lift                         & 98.5           & 99.0        & 98.5                              \\
Can                          & 61.0             & 68.0        & 70.0                               \\
Square                       & 35.0            & 47.5       & 48.5                                               \\ \bottomrule
\end{tabular}
}
    \caption{Quantitative Ablation of the effect of the number of 3D Gaussians on the performance of the BCQ algorithm. }
    \label{tab: ablation about the point number}
    \vspace{-6mm}
\end{table}

\subsection{Experimental Settings}
This work proposes the effectiveness of the novel scene representation on RL. Therefore, we compare the performance of different modalities on different RL algorithms. For each environment, we set several cameras to observe the scenario including the agentview camera and the robot hand and shoulder cameras. We set the baselines as the other three explicit modalities, namely multiview image, point cloud, and voxels for all algorithms.
We exclude the implicit NeRF representation because it cannot be generalized to different scenes thereby needing to retrain the model on each task, and requires additional masks.
The multiview images denote that we stack these images along the channel dimension and input to the RL network. We choose a simple ResNet18 as the encoder. The point cloud is produced by these ground-truth depth maps. Both points and 3DGS utilize the SetTransformer as the encoder. We set the voxel grid as $V \in R^{64 \times 3}$, and trilinearly interpolate the point clouds into it. We adopt a 3D convolutional Resnet to encode it. We set the batch size of voxel representation as 32, but that of the other modalities is 100. This is because 3D Convolution could consume a lot of memories. We train all models on an NVIDIA A6000 GPU with the Adam optimizer. 
It is noted that our goal is not to maximize the performance of each RL algorithm on specific tasks. In contrast, we aim to demonstrate the scene representation is general and effective. Therefore we do not search for the optimal hyperparameters for every task. 
As stated in Fig.~\ref{fig:main-fig}, after the Generalizable Gaussian model is trained, we fixed it to be an encoder to transfer multiview image observations into the 3D-consistent and geometry-aware Gaussian representation. The RL network takes the Gaussians as input to predict proper actions and then obtain the next state. The new state produces new observations which will be iteratively transformed into the 3DGS representation in the next loop. 
\subsection{Results Analysis}
We report all quantitative comparisons across different modalities and offline algorithms in Table~\ref{tab: result 1}. The success rates are evaluated with randomly initialized shapes and positions in the online simulation environment. It is observed that our model delivers the most satisfactory results in most cases. In contrast, the cases in which the point cloud representation is only used show under-average performance compared with the other benchmarks. This indicates that these extra Gaussian properties improve the ability to represent the scene on account of detailed local geometry descriptions. In Lift, Square, and Transport tasks, our representation outperforms all counterparts across all three baseline algorithms. Our performance in the Square environment for BCQ method improves 15$\%$, 70$\%$, and  15$\%$ compared with the other two modalities. In the most difficult scenario Transport, BCQ and IQL algorithms fully collapse for all modalities due to the nature of those methods. But our method is $10\%$, $44\%$, and $15\%$ better than the rest three representations in IRIS experiments. Besides, the covariance of our success rate is relatively smaller than others, which indicates that our method is more stable. 
\begin{table}[t]
    \centering
\scalebox{1.2}{
\begin{tabular}{ccccc}
\toprule
Training step                 & PSNR & Lift & Can & Square\\ \midrule
3500                         & 25.9           & 97.5        & 70.5     & 42.0                             \\
9500                          & 28.8             & 97.5        & 71.0       & 43.0                             \\
15000                       & 31.5            & 97.0       & 71.5    & 44.5 
\\
25000                       & 33.7            & 98.5       & 70.0    & 48.5
\\ \bottomrule
\end{tabular}
}
    \caption{Quantitative Ablation study to show the effect of 3DGS rendering quality on the performance. }
    \label{tab: ablation about the PSNR}
    \vspace{-5mm}
\end{table}
We additionally evaluate the influence of the number of points on our performance. We downsample the original point set to 8192, 4096, and 2048 points respectively, and test them on the BCQ algorithm. The results are shown in Table~\ref{tab: ablation about the point number}. The metrics in this Table are the average success rate. Notably, the proposed approach is overall not sensitive to the number of points. The performance is likely to be improved further as the point number increases, but the trend is not obvious. Interestingly, 4096 points performed just as well as 8192 points, and sometimes even better. On the contrary, when the number of points decreases to 2048, the simple task Lift is unaffected but the performance on relatively hard tasks, i.e. Can and Square, drops by 12$\%$ and 26$\%$ respectively. Therefore, this should be a trade-off to determine how many points are used in scene representation, and harder tasks naturally require more points to perceive. 
Furthermore, we test the effect of the quality of 3DGS reconstruction on RL performance. We stopped the GS model training to obtain models with different reconstruction qualities. Then we test each model on the BCQ algorithm and the report is listed in Table~\ref{tab: ablation about the PSNR}. The rendering quality, depicted by PSNR, indeed affects the RL performance, especially on relatively harder tasks. More precise reconstruction leads to better RL performance.

According to this point, we, in addition, evaluate the effectiveness of some basic designs in the generalizable GS framework and report their results in Table~\ref{tab: ablation of main components}. We use PSNR, SSIM \cite{wang2004image}, and vgg-based LPIPS \cite{zhang2018unreasonable} to evaluate the reconstruction quality. In this Table, w/o feat$\_$inp means the regressor module does not reuse the pre-extracted features ($F_{s1}, F_{s2}$ in Eq.~\ref{eq : image feature extraction}), it replaces features with images. w/o refine denotes that we remove the Gaussian refinement module. "prediction depth" and "real depth" are the depth maps we use in subsequent modules because the Robomimic can also provide the real depth map. It can be seen that both cascaded architecture in feature space and the Gaussian refinement are effective in improving the reconstruction quality. Moreover, even though we replace the input depth map with the real one, there is only a minimal improvement, which indicates that the proposed approach can work well whether the observation includes depth information or not. 


\begin{table}[t]
    \centering
\scalebox{1.2}{
\begin{tabular}{cccc}
\toprule
                 & PSNR & SSIM & LPIPS \\ \midrule
w/o feat\_inp                   & 31.62             & 0.979        & 0.081       
                                \\
w/o refine                         & 32.24           & 0.971        & 0.062                               \\
prediction depth                       & 33.69            & 0.985       & 0.041    
\\
real depth                       & 33.91            & 0.987       & 0.038   
\\ \bottomrule
\end{tabular}
}
    \caption{Ablations of some main components design. }
    \label{tab: ablation of main components}
    \vspace{-6mm}
\end{table}
\section{Conclusion}
\label{sec:conclusion}
We propose to use a novel scene representation i.e. 3DGS, for vision-based RL algorithms. Traditional 3DGS is not suited to be an environment representation due to the need for per-scene optimization. To address this, we introduce the learning framework that directly predicts 3D Gaussians from visual observations in a generalizable manner, which contains three main procedures, a depth estimator, a Gaussian regressor, and a Gaussian refinement module. This framework can effectively capture task-specific priors of image-to-local structures. To evaluate the effectiveness of the novel scene representation, we compare it with other explicit vision representations on the Robomimic platform with four different tasks and across three different RL algorithms. The results show that our generalizable GS representation overall outperforms these counterparts, and improves the Success Rate by 10$\%$, 44$\%$, and 15$\%$ respectively on the most challenging task.


\bibliographystyle{IEEEtran}
\typeout{}
\bibliography{IEEEabrv,mybibfiles}
\end{document}